\documentclass[a4paper]{article}
\usepackage{aaai_custom}
\usepackage[square,numbers]{natbib}

\usepackage{times}  
\usepackage{helvet}  
\usepackage{courier}  
\usepackage{url}  
\usepackage{graphicx}  
\usepackage{hyperref}
\newcommand{\todo}[1]{}
\renewcommand{\todo}[1]{{\color{red} TODO: {#1}}}

\title{Injecting and removing malignant features in mammography with
CycleGAN: Investigation of an automated adversarial attack using
neural networks}

\author{\AND
Anton S.~Becker\thanks{Correspondence to: Anton S.~Becker, Institute of Diagnostic and Interventional Radiology, University Hospital Zurich, Raemistrasse 100, CH-8091 Zurich, E-Mail: \url{anton.becker@usz.ch}}~\textsuperscript{1,2)}\\
\And 
Lukas Jendele\thanks{Joint contribution}~\textsuperscript{3)}\\
\And
Ondrej Skopek\textsuperscript{$\dagger$~3)}\\
\AND
Nicole Berger~\textsuperscript{1)}
\And
Soleen Ghafoor~\textsuperscript{1,4)}\\
\And
Magda Marcon~\textsuperscript{1)}\\
\And
Ender Konukoglu~\textsuperscript{5)}\\
\AND \vspace{0.001cm}\\
\textsuperscript{1)} Institute of Diagnostic and Interventional Radiology, University Hospital of Zurich\\
\textsuperscript{2)} Department of Health Sciences and Technology, ETH Zurich\\
\textsuperscript{3)} Department of Computer Science, ETH Zurich\\
\textsuperscript{4)} Department of Radiology, Memorial Sloan Kettering Cancer Center, New York City\\
\textsuperscript{5)} Computer Vision Laboratory, Department of Information Technology and Electrical Engineering, ETH Zurich%
}

\begin{document}
\nocopyright
\maketitle

\begin{abstract}
\textbf{Purpose}\quad To train a cycle-consistent generative adversarial network (CycleGAN) on mammographic data to inject or remove features of malignancy, and to determine whether
these AI-mediated attacks can be detected by radiologists.\\
\textbf{Material and Methods}\quad From the two publicly available datasets, BCDR and 
INbreast, we selected images from cancer patients and healthy controls. An internal dataset 
served as test data, withheld during training. We ran two experiments training CycleGAN on 
low and higher resolution images ($256 \times 256$ px and $512 \times 408$ px). Three 
radiologists read the images and rated the likelihood of malignancy on a scale from 1--5
and the likelihood of the image being manipulated. The readout was evaluated by ROC
analysis (Area under the ROC curve = AUC).\\
\textbf{Results}\quad At the lower resolution, only one radiologist exhibited markedly
lower detection of cancer (AUC=0.85 vs 0.63, p=0.06), while the other two were unaffected 
(0.67 vs. 0.69 and 0.75 vs. 0.77, p=0.55). Only one radiologist could discriminate between 
original and modified images slightly better than guessing/chance (0.66, p=0.008). At the 
higher resolution, all radiologists showed significantly lower detection rate of cancer in 
the modified images (0.77--0.84 vs. 0.59--0.69, p=0.008), however, they were now able to
reliably detect modified images due to better visibility of artifacts (0.92, 0.92 and 0.97). \\
\textbf{Conclusion}\quad
A CycleGAN can implicitly learn malignant features and inject or remove them
so that a substantial proportion of small mammographic images would consequently be
misdiagnosed. At higher resolutions, however, the method is currently limited and has a
clear trade-off between manipulation of images and introduction of artifacts.
\end{abstract}

\section{Introduction}
Machine learning (ML) in medical imaging is a promising field of research, which will bring
substantial changes to radiology in the coming years. Mammography, as a 2D x-ray
projection modality with great clinical significance, is arguably one of the first fields where
these techniques will be successfully deployed. Many ML studies focus on (semi) automated
detection \citep{Becker:2017de} or classification of cancer \citep{Truhn:2018dn}. However, there is also a considerable number
of studies focusing on radiation dose reduction e.g. by reconstructing images from the raw
data through ML algorithms \citep{Chen:2017iy} or working directly with ultralow-dose data \citep{Schwyzer:2018gw}. In mammography, this would be of great benefit for all
patients, and in particular for young women who are more vulnerable to the effects of
ionizing radiation. First ML algorithms in CT are already applied in the clinical routine,
autonomously optimizing patient table positioning and thus reducing the applied radiation
dose \citep{Saltybaeva:2018gm}. Most advanced ML algorithms are fundamentally opaque and as they,
inevitably, find their way onto medical imaging devices and clinical workstations, we need
to be aware that they may also be used to manipulate raw data and enable new ways of
cyber-attacks, possibly harming patients and disrupting clinical imaging service \citep{Mahler:2018uf}.

One specific genre of ML algorithms, Generative Adversarial Networks (GANs), are of
particular importance in this context. GANs are a subclass of deep learning algorithms, itself
a class of algorithms within the realm of ML or artificial intelligence (AI) \citep{LeCun:2015dta}. A GAN consists
of two neural networks competing against each other: The first, generator network (G),
manipulates sample images and the second, discriminator network (D), has to distinguish
between real and manipulated samples \citep{Goodfellow:2014wp}. Due to their opposed cost function, the neural
networks are competing against each other in order to improve their performance (in game
theory this scenario is known as a ``two person zero-sum game'' \citep{Myerson:1997vw}). Given infinite
resources and time, this will theoretically result in G producing samples from the real image
distribution (i.e. perfect manipulations) and D completely incapable of discriminating,
giving each such sample a probability of 0.5 for being either manipulated or real. In our
case, we hypothesized that a GAN can learn an implicit representation of what cancer in
mammography looks like, and specifically alter images, so they would be misdiagnosed
(healthy as cancerous and vice versa) while even a radiologist could not differentiate
between manipulated images and real ones.

Hence, the purpose of this study was to train a pair of GANs on mammographic data to
inject or remove features of malignancy and to determine whether these AI-mediated
attacks can be detected by radiologists.

\section{Methods}
\subsection{Patient Cohorts/Datasets}
From two publicly available datasets, BCDR \citep{Moura:2013jo} and INbreast \citep{Moreira:2012gj}, 680 mammographic
images from 334 patients were selected, 318 of which exhibited potentially cancerous
masses, and 362 were healthy controls. We used all INBREAST cases with BI-RADS 3 or
greater as cancer cases, and all cases with a focal lesion and marked as ``malignant'' from
BCDR. As an additional test dataset for experiment two (see below), we used images from a
private dataset previously published in \citep{Becker:2017de} – (302 cancer / 590 healthy). These images
were withheld from the network during training and only used to generate images for the
readout and test how well the network generalizes to new, unseen images.

\subsection{GAN Model Selection and Adaptation}
We view the task of injecting and removing malignant features from an image as an image
translation problem in the spirit of the recently proposed cycle-consistent GANs model
(CycleGAN) \citep{Zhu:2017ua}, which aim to translate images from one distribution, e.g. healthy subjects,
to another distribution, e.g. cancer patients, and back. We trained an adapted version of
CycleGAN, using two pairs of generator and discriminator networks to convert cancerous
breast images to healthy and back, and vice versa for the controls.

\subsubsection{First Experiment}
The CycleGAN architecture was implemented in TensorFlow v1.5 \citep{tensorflow}. Images were rescaled to $256 \times 256$ px, normalized between $-1$ and $+1$, and augmented tenfold by random rotation, scaling, and contrast perturbations. The training was performed on a consumer-grade personal computer (PC) with an Nvidia GeForce GTX 1070 graphics processing unit (GPU). The code and toy data for the first experiment can be found online: \href{https://github.com/BreastGAN/experiment1}{\textit{github.com/BreastGAN/experiment1}}. It contains all the relevant hyperparameters and was designed to run out-of-the-box via Docker to facilitate reproduction and extension of our results.

\subsubsection{Second Experiment}
This experiment was designed and conducted after the first readout in order to further test the limits of CycleGAN. We increased the resolution of the images more than 3-fold to $512 \times 408$ px. After an initial test-run with satisfactory results, we decided to proceed without data augmentation. Due to the increased image size, we used a GPU cluster consisting of up to eight GeForce GTX TITAN X/Xp GPUs. We implemented CycleGAN in Tensorflow v1.12.rc2 for this experiment. The code and synthetic data for the second experiment can be found online: \href{https://github.com/BreastGAN/experiment2}{\textit{github.com/BreastGAN/experiment2}}.

\subsection{Radiologist readout}
\subsubsection{First readout}
From the first experiment, we randomly chose 30 modified and 30 original images, with 40 images in pairs and 20 unpaired images (cancer vs. healthy). Only images with visible masses at this resolution were considered from the original images from the respective category. The images were presented to three radiologists (5 years of experience for the two senior readers, and one PGY-6 fellow) who rated them on a 5-point Likert-like scale for the likelihood of cancer (``how likely would you recall this patient'') and had to indicate whether the image was real or generated/modified. In the first readout, this was a binary indication. The radiologists were fully blinded to the purpose of the study and the distribution of cancer vs. healthy cases, i.e. they were only informed that some images had been modified ``by the computer''. Reference standard was the original label/class of the image also in the GAN-modified images, matching our scenario of the images being modified in the background to ``fool'' the radiologist.

\subsubsection{Second Readout}
In the second readout, the readers knew the results of the initial readout, roughly what types of artifacts were to be expected, and that CycleGAN was used in the study. During training we observed that artifacts seemed to get more pronounced in the later training process (see results section below). To test this hypothesis, we presented modified images generated after different number of training iterations (35k and 70k) and let the readers rate the artificial artifacts on a 5-point scale as well.

They were again blinded to the distribution of samples: From step 35k we selected 6 image pairs and 6 single images (evaluation dataset used during training) to test for differences in artifact occurrence; after training the GAN for 70k iterations we selected 12 healthy and 12 cancerous images. Half of the images were modified and half of them were originals, and again half of them paired and the other half unpaired, from each the evaluation and the test dataset. Hence, the total number of images for the second readout was 72 (36 cancer / 36 healthy). 

\subsection{Statistical Analysis}
Statistical analysis was performed using R v.3.4.4. (R Foundation for Statistical Computing, Vienna, Austria). Continuous data was expressed as median and interquartile range (IQR). Categorical data was given in absolute counts.

Detection accuracy was assessed with receiver operating characteristic (ROC) analysis. ROC curves were computed with the package pROC v.1.12.1, the discriminatory performance of readers was expressed as the area under the ROC curve (AUC). AUC were compared with DeLong's non-parametric test \citep{delong1988comparing}. Where helpful, p-values of the three readers were combined with the procedure proposed by Stouffer et al.~\citep{stouffer1949american}.

\section{Results}
\subsection{First experiment}
In a first experiment, we modified CycleGAN \citep{Zhu:2017ua} to work with small mammographic images ($256 \times 256$ px) from the publicly available datasets BCDR \citep{Moura:2013jo} and INBreast \citep{Moreira:2012gj}, running on a consumer grade PC. 

Qualitatively, we noticed that at the beginning of the training, during the initial iterations, the GAN started out by first adjusting global features like contrast/brightness and then started removing or adding glandular tissue early on, thus increasing the overall breast density. Later, it would pick up skin-thickening as a malignant feature. Finally, it would apply more focal alterations like removing or adding mass-like lesions, or morphing large, benign calcifications into fat or soft tissue masses. In general, poorly circumscribed, malignant looking masses would be preferentially placed on top of preexisting structures (either islets of breast tissue or benign findings). Moreover, added focal lesions generally looked somewhat more realistic than removed ones. We noticed that after 160k steps, grid-like or checkerboard-like artifacts became very prominent in the generated images, making it fairly easy for humans to spot the manipulated images. Hence, we went back to check the images for less pronounced artifacts and loaded the network with weights before step 160k to generate the images for the first readout. The code to reproduce this first experiment can be found on \href{https://github.com/BreastGAN/experiment1}{\textit{github.com/BreastGAN/experiment1}}.

\begin{figure*}[t]
    \centering
    \includegraphics[width=\linewidth]{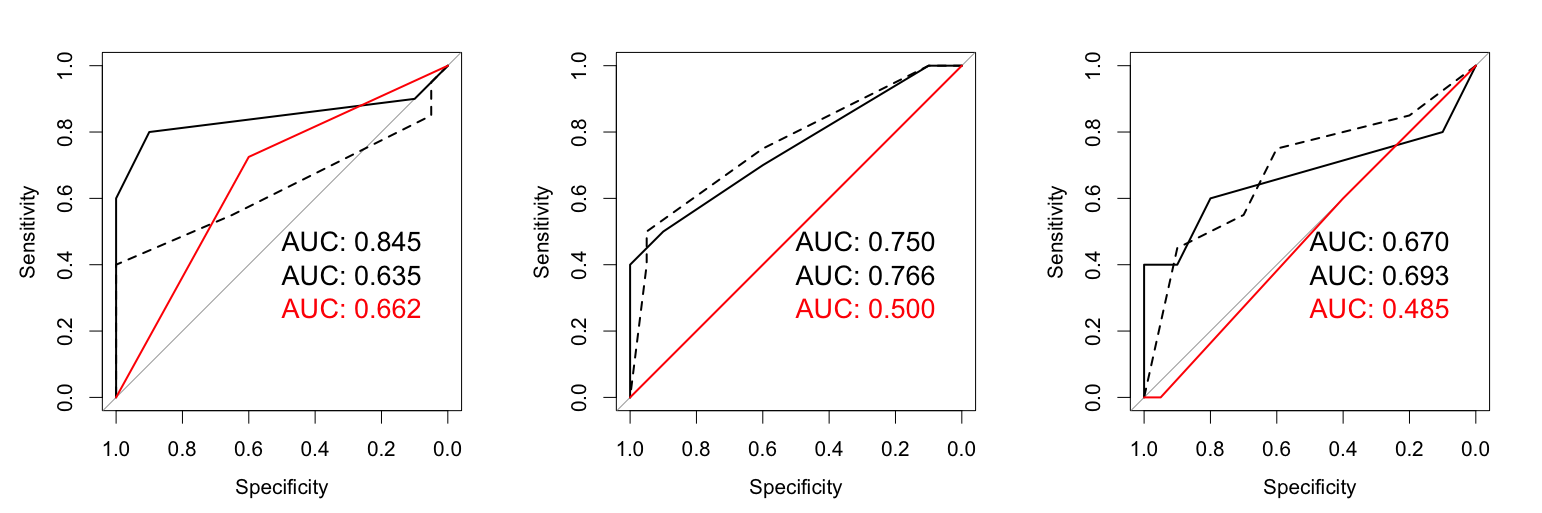}
    \caption{ROC curves for cancer detection in the original images (solid line), the modified images (dashed line) and the distinction between original and modified image (red line) for all three readers in the first readout (low resolution, $256 \times 256$ pixels).}
    \label{fig:exp1}
\end{figure*}

From the images generated by the network trained less than 160k steps, we randomly chose 30 modified and 30 original images, with 40 images in pairs (i.e. the original and GAN-modified version), and 20 unpaired images. These images were presented to 3 radiologists (in random order), who had to rate the likelihood of malignancy on a scale from 1--5 and indicate whether the image was modified or not (binary decision). We found that in one of the experienced radiologists, the modifications introduced by CycleGAN markedly reduced diagnostic performance. The AUC of this reader dropped from 0.85 to 0.63 (p=0.06) in the modified images, with regard to the original labels/classes, while the two other readers seemed unaffected, however, at a lower baseline performance (AUC 0.75 vs. 0.77 and 0.67 vs. 0.69, p=0.55). Only the first reader could detect the CycleGAN modifications in some images (AUC=0.66, p=0.008), whereas the two other readers were not better than chance in this task (AUC=0.48 and 0.50, p=0.59 and 0.50). These results are summarized in Figure~\ref{fig:exp1}.

\subsection{Second experiment}
In the second experiment, we further investigated the occurrence of artifacts at later training stages and in images with a higher resolution. Since a higher resolution allows for finer textures and more detail, which are essential in the clinical use of mammographies, we hypothesized that the artifacts would be easier to spot. Therefore, we more than tripled the working resolution of our CycleGAN to $512\times408$ px.  Due to the increased memory demand, we ran our experiments on a dedicated GPU cluster. The source code together with toy data is available online on \href{https://github.com/BreastGAN/experiment2}{\textit{github.com/BreastGAN/experiment2}}. To test how well the network generalized to new, unseen data, we used an additional, internal test dataset from a prior study \citep{Becker:2017de}, which was withheld during training.

\begin{figure*}
    \centering
    \includegraphics[]{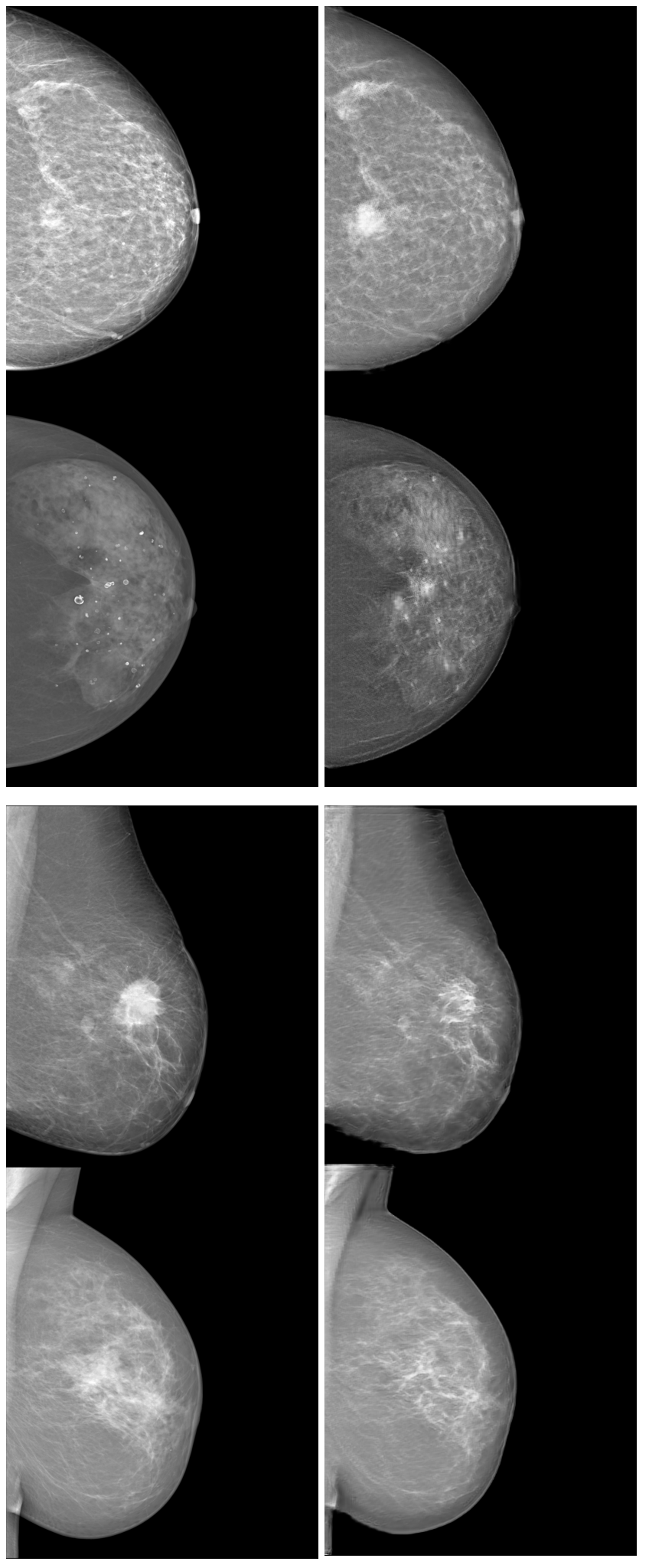}
    \caption{Representative examples of original (left) and CycleGAN-modified images (right) from the second experiment. The top two rows are healthy mammographies with injected malignancies, the bottom two rows are images with real cancers, which were removed by CycleGAN. Note how in the top examples (healthy to cancer), the GAN uses existing features in the image to modify in order to look malignant (islet of breast tissue and benign macrocalcifications, respectively).}
    \label{fig:example}
\end{figure*}

On inspection of the training monitoring, we noticed the same learning pattern as the first time, however, the grid-like artifacts were indeed more pronounced and seemed to increase after around 45--50k steps of training iteration. Hence, we selected 6 image pairs and 6 images from step 35k (evaluation dataset used during training) to test for differences in artifact occurrence. From step 70k we selected 12 healthy and 12 cancerous images, half of them modified and half of them originals, and half of them paired and the other half unpaired, from each the evaluation and the test dataset. The total number of images for the second readout was 72 (36 healthy / 36 cancer), a representative selection of images is shown in Figure~\ref{fig:example}.

\begin{figure*}[t]
    \centering
    \includegraphics[width=\linewidth]{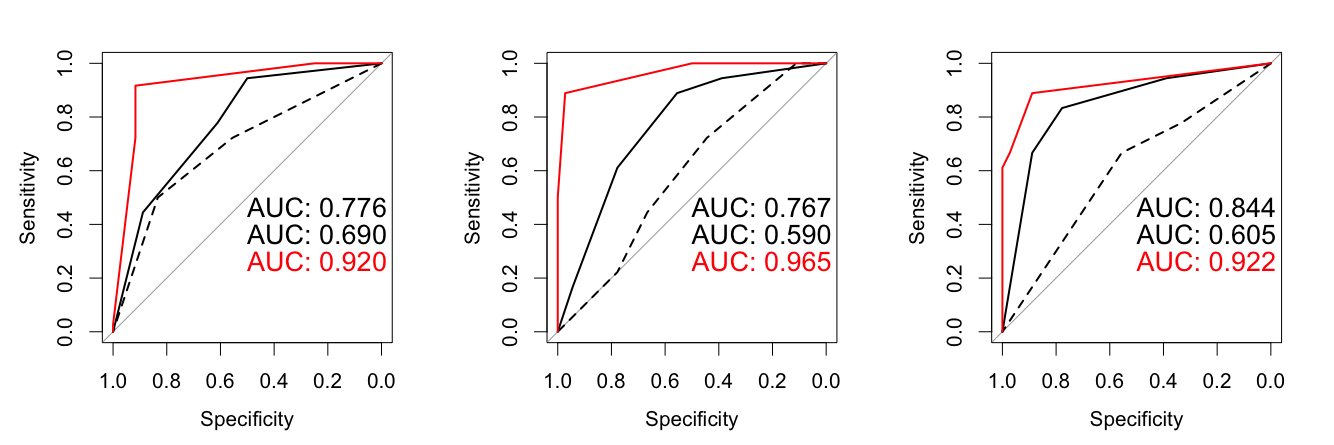}
    \caption{ROC curves for cancer detection in the original images (solid line), the modified images (dashed line) and the distinction between original and modified image (red line) in all three readers for the second readout (higher resolution, $512 \times 408$ pixels). The significantly lower performance in modified images can be clearly appreciated, albeit these were easily identified as such, as apparent by the red curve.}
    \label{fig:exp2}
\end{figure*}

We found that for all radiologists, the performance to discriminate between healthy and cancerous (referencing the original image class) dropped significantly (AUCs given in Table \ref{tab:tab} and Figure~\ref{fig:exp2}, p=0.008), with no significant difference between evaluation and test data (p=0.53--0.80), indicating that the network generalized well to new data. However, all radiologists could now reliably identify the modified images (AUC=0.97--0.92), confirming our hypothesis that the artifacts were easier to identify at higher resolution. Identification of modifications was not better in images from the later training stages (AUC early vs. late = 1.0 vs. 0.92, 0.91 vs. 0.94 and 0.93 vs. 0.93), which did not confirm our hypothesis that the GAN would produce more artifacts at later training stages. 

\begin{table*}[t]
    \centering
    \begin{tabular}{l|lll}
         \textbf{READER} & \textbf{AUC originals} & \textbf{AUC modified} & \textbf{\textit{p}-value} \\
        \hline
         1 &  0.78 & 0.69 & 0.12 \\
         2 &  0.77 & 0.59 & 0.10 \\
         3 &  0.84 & 0.60 & 0.02 
    \vspace{0.25cm} \\
         \textbf{READER} & \textbf{AUC evaluation} & \textbf{AUC test} & \textbf{\textit{p}-value} \\
        \hline
         1 &  0.60 & 0.56 & 0.74 \\
         2 &  0.39 & 0.43 & 0.80 \\
         3 &  0.65 & 0.56 & 0.52
    \vspace{0.25cm} \\    
         \textbf{READER} & \textbf{AUC late} & \textbf{AUC early} & \textbf{\textit{p}-value} \\
        \hline
         1 &  0.93 & 0.93 & 1.00 \\
         2 &  1.00 & 0.93 & 0.16 \\
         3 &  0.90 & 0.95 & 0.52 \\
    \end{tabular}
        
    \caption{Results of the ROC analysis for the second experiment. Combined p-values were 0.008 for original vs. modified, 0.81 for evaluation vs. test set and 0.79 for late vs. early training stage.}
    \label{tab:tab}
\end{table*}

\section{Discussion}
In the present study, we investigated whether a GAN can inject or remove malignant features in a realistic way that would make modified images indistinguishable from real ones even for radiologists. We found that at low and slightly higher resolution, these features were realistic enough to change the radiologist’s diagnosis from healthy to suspicious and vice versa. While at low resolution, there were no or very little artifacts that distinguished the modified images from real ones, at the higher resolution these artifacts became obvious to the point that the modified images were recognized more easily. We did not find any difference in appearance between training and test data, suggesting that the GAN generalizes well to unseen data.

In the past decades, computers have become an integral part of all clinical workflows in modern healthcare systems. This brings great advantages on one hand, i.e. better documentation, more efficient workflows \citep{Yala:2016ez} or new discoveries in research \citep{Chan:gu}, but on the other hand, the system becomes dependent on computers and will inevitably take over some of their inherent weaknesses. The fact that such scenarios are beyond hypothetical deliberations has painfully become apparent in the recent cases where patient information in whole hospital systems was encrypted and thus rendered inaccessible by attackers who demanded a ransom payment for decryption (this particular kind of attack is called ``ransomware'' \citep{Kharraz:2015gb}. Moreover, these potential threats are clearly not only limited to healthcare: For example, a government investigation concluded that in the past 2016 U.S. presidential election, cyber warfare may have played a role in swaying the results in favor of a particular candidate \citep{Dir_Intelligence:2017wo}.

All modalities in a modern medical imaging department rely heavily on computers and networks, making them a prime target for cyber-attacks \citep{Mahler:2018uf}. As machine learning or artificial intelligence (AI) algorithms will increasingly be used in the clinical routine, whether to reduce the radiation burden by reconstructing images from low-dose raw data \citep{Chen:2017iy, Schwyzer:2018gw} or help diagnose diseases \citep{Becker:2017de, Becker:2017ji, Rajpurkar:2017tu, DeFauw:2018fc} their widespread implantation would also render them attractive targets for attacks. Exploiting vulnerabilities of deep neural networks is becoming an established field of research, yielding interesting results like the ``one-pixel attack'' \citep{Su:2017uw}, where an attacking neural network only modifies one pixel in order for the image to be misclassified. Evidently though, such an attack would not be able to fool a human observer. Hence, an important aspect of GANs is that they may be able to produce realistic examples which could mislead human observers as well as machine algorithms \citep{Finlayson:2018ti}. Regarding medical imaging, we can imagine two categories of attacks: Focused and generalized attacks. In a focused attack, an algorithm would be altered so it would misdiagnose a targeted person (e.g. political candidate or company executive) in order to achieve a certain goal (e.g. manipulation of election or hostile company takeover). In a generalized attack, a great number of devices would be infected with the malicious algorithm lying dormant most of the time and stochastically leading to a certain number of misdiagnoses, causing potentially fatal outcomes for the affected patients, increased cost for the whole healthcare system and --- ultimately --- undermining the public's trust in the healthcare system. At the time of writing, however, we would argue that the technology is not yet advanced enough to make the threat of such an attack imminent. However, we think this matter deserves attention and further investigation in order to secure software/algorithms and hardware, before technology catches up.

It is worth pointing out that there are also many other possible applications of GANs apart from cyber-attacks. In a recent study, the authors investigated which features are learned by a GAN when estimating the severity of congestive heart failure in a chest x-ray examination \citep{Seah:2018ej}. Hence, GANs could be used either to discover new imaging features of a disease, for teaching purposes, or to detect biases and confounders in training datasets. Furthermore, many datasets, especially in a screening setting, are highly unbalanced, i.e. the cases of healthy individuals far outweigh the cancer cases. GANs could be used to create more balanced datasets and thus facilitate training of other ML algorithms.

There are several limitations that need to be mentioned. The introduction of grid or checkerboard artifacts is a known problem in GANs related to upsampling \citep{Odena:2016fy}. We attribute the more perceptible artifacts at the higher resolution to two reasons: First, the higher resolution allows for finer textures and details, and thus will require more careful modifications by the GAN in order not to distort the natural patterns occurring in the fatty and dense breast tissue. Second, although we combined two of the largest publicly available datasets, they are still fairly small compared with datasets currently used in computer vision research: For example, the ImageNet database contains nearly 14.2 million images at the time of writing \citep{ImageNet:we}. Moreover, the average resolution of ImageNet pictures is lower than the resolution used in our second experiment. This entails a relative sparsity of the dataset in our experiments for the task at hand \citep{Zhu:2017ua} leading to overfitting and artifacts \citep{Odena:2016fy}. Both points are highly relevant for mammography: Clinical mammographic images have a very high resolution, about two orders of magnitude higher than our experiments. For future research, much larger databases with mammographic images will be needed.

Increasing the size of the images brings another problem about: One of the most important bottlenecks for deep learning experiments in computer vision is memory. Tackling this problem is non-trivial and an active field of research \citep{Kennedy_online:2017gpu}. Hence, it is currently common practice in research to resize the images to a low resolution to speed up the training process or make it feasible at all. We were left to choose between resizing the images thus losing detail information or working with small patches of full-resolution mammographies while losing global information. Since this was a proof-of-principle study and we were interested in whether, how and where a GAN would extract and insert features in the whole image, we chose the former trade-off. Moreover, a readout with single image patch is less representative of clinical routine than small, resized versions of the whole mammography.

In conclusion, we could show that a CycleGAN is capable of implicitly learning malignant features and injecting or removing them so that a substantial proportion of small mammographic images are consequently misdiagnosed. At higher resolutions, however, the method is limited and currently has a clear trade-off between manipulation of images and introduction of artifacts. Nevertheless, this matter deserves further study in order to shield future devices and software from AI-mediated attacks.

\bibliography{doc}
\bibliographystyle{unsrt}

\end{document}